\documentclass[runningheads]{llncs}
\usepackage{graphicx}

\usepackage{tikz}
\usepackage{comment}
\usepackage{amsmath,amssymb} 
\usepackage{color}

\usepackage{epsfig}
\usepackage{graphicx}
\usepackage{cite}
\usepackage{booktabs}
\usepackage{subfig}

\usepackage{wrapfig}
\usepackage{caption}

\usepackage{multirow}
\usepackage{verbatim}
\usepackage{hyperref}
\hypersetup{pagebackref,breaklinks,colorlinks}

\usepackage[accsupp]{axessibility}  

\newcommand{\etal}{\textit{et al.}}

\newcommand{\eg}{\textit{e.g.}}

\usepackage[capitalize]{cleveref}
\crefname{section}{Sec.}{Secs.}
\Crefname{section}{Section}{Sections}
\Crefname{table}{Table}{Tables}
\crefname{table}{Tab.}{Tabs.}

\begin{document}
\pagestyle{headings}
\mainmatter
\def\ECCVSubNumber{4247}  

\title{Depth Map Decomposition for Monocular Depth Estimation}

\titlerunning{Depth Map Decomposition for Monocular Depth Estimation}
%
\author{Jinyoung Jun\inst{1}\orcidID{0000-0002-8256-6846} \and
Jae-Han Lee\inst{2}\orcidID{0000-0002-3674-4023} \and \\
Chul Lee\inst{3}\orcidID{0000-0001-9329-7365} \and
Chang-Su Kim\inst{1}\orcidID{0000-0002-4276-1831}}
\authorrunning{J. Jun et al.}

\institute{School of Electrical Engineering, Korea University, Seoul, Korea \and
Gauss Labs Inc. \and
Department of Multimedia Engineering, Dongguk University, Seoul, Korea\\
\email{jyjun@mcl.korea.ac.kr, jaehan.lee@gausslabs.ai,\\ chullee@dongguk.edu, changsukim@korea.ac.kr}}
\maketitle

\begin{abstract}
We propose a novel algorithm for monocular depth estimation that decomposes a metric depth map into a normalized depth map and scale features. The proposed network is composed of a shared encoder and three decoders, called G-Net, N-Net, and M-Net, which estimate gradient maps, a normalized depth map, and a metric depth map, respectively. M-Net learns to estimate metric depths more accurately using relative depth features extracted by G-Net and N-Net. The proposed algorithm has the advantage that it can use datasets without metric depth labels to improve the performance of metric depth estimation. Experimental results on various datasets demonstrate that the proposed algorithm not only provides competitive performance to state-of-the-art algorithms but also yields acceptable results even when only a small amount of metric depth data is available for its training.

\keywords{Monocular depth estimation, relative depth estimation, depth map decomposition}
\end{abstract}

\section{Introduction}
\label{sec:introduction}
Monocular depth estimation is a task to predict a pixel-wise depth map from a single image to understand the 3D geometry of a scene. The distance from a scene point to the camera provides essential information in various applications, including 2D-to-3D image/video conversion \cite{Xie2016ECCV}, augmented reality \cite{Liu2018CVPR}, autonomous driving \cite{Godard2017CVPR}, surveillance \cite{kim2016weighted}, and 3D CAD model generation \cite{Izadinia2017CVPR}. Since only a single camera is available in many applications, monocular depth estimation, which infers the 3D information of a scene without additional equipment, has become an important research topic.

Recently, learning-based monocular depth estimators using convolutional neural networks (CNNs) have shown significant performance improvements, overcoming the intrinsic ill-posedness of monocular depth estimation by exploiting a huge amount of training data \cite{eigen2014depth, laina2016deeper, xu2017multi, heo2018monocular, fu2018deep, chen2019structure, yin2019enforcing, lee2020multi, bhat2021adabins}. Existing learning-based monocular depth estimators can be classified into two categories according to the properties of estimated depth maps: relative depth estimation and metric depth estimation. Relative depth estimation predicts the relative depth order among pixels \cite{zoran2015learning, chen2016single, xian2020structure, lienen2021monocular}. Metric depth estimation, on the other hand, predicts the absolute distance of each scene point from the camera \cite{eigen2014depth, laina2016deeper, xu2017multi, heo2018monocular, chen2019structure}, which is a pixel-wise regression problem.

\begin{figure}[!t]
  \centering
   \includegraphics[width=\linewidth]{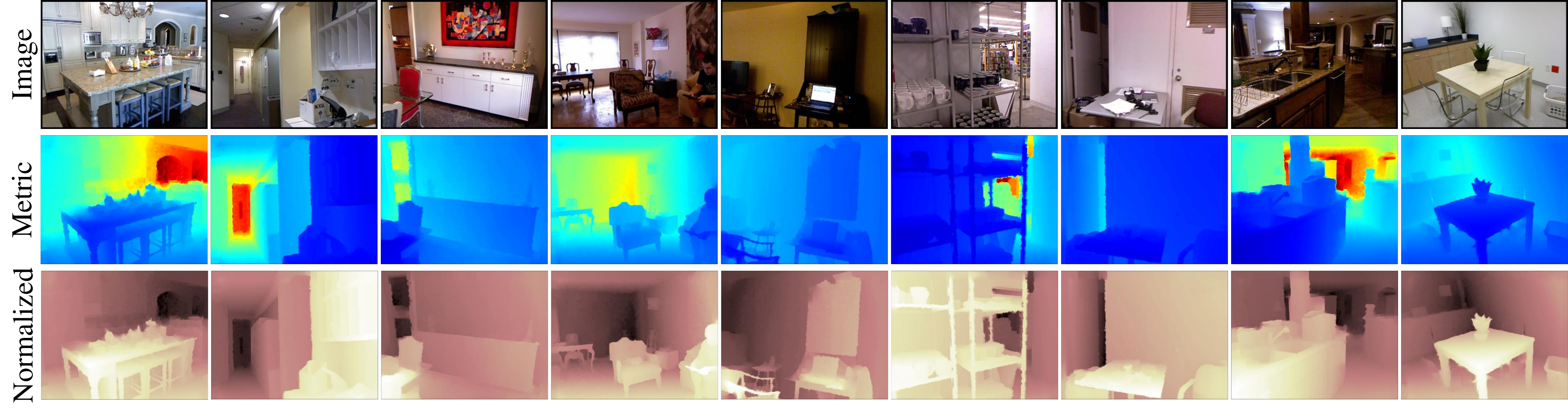}
   \caption{Metric depth maps and their normalized depth maps.}
   \label{fig:intro}
\end{figure}

To estimate a metric depth map, a network should learn both the 3D geometry of the scene and the camera parameters. This implies that a metric depth estimator should be trained with a dataset obtained by a specific camera. In contrast, a relative depth estimator can be trained with heterogeneous datasets, \eg, disparity maps from stereo image pairs or even manually labeled pixel pairs. Thus, relative depth estimation is an easier task than metric depth estimation is. Moreover, note that the geometry of a scene can be easily estimated when extra cues are available. For example, depth completion \cite{ma2018sparse, xu2019depth, park2020non}, which recovers a dense depth map from sparse depth measurements, can be performed more accurately and more reliably than monocular depth estimation is. Based on these observations, metric depth estimation algorithms using relative depths as extra cues have been developed via fitting \cite{ranftl2020, lienen2021monocular} or fine-tuning \cite{ranftl2021vision}.

In this paper, we propose a monocular metric depth estimator that decomposes a metric depth map into a normalized depth map and scale features. As illustrated in Fig.~\ref{fig:intro}, a normalized depth map contains relative depth information, and it is less sensitive to scale variations or camera parameters than a metric depth map is. The proposed algorithm consists of a single shared encoder and three decoders, G-Net, N-Net, and M-Net, which estimate gradient maps, a normalized depth map, and a metric depth map, respectively. M-Net learns to estimate metric depth maps using relative depth features extracted by G-Net and N-Net. To this end, we progressively transfer features from G-Net to N-Net and then from N-Net to M-Net. In addition, we develop the mean depth residual (MDR) block for M-Net to utilize N-Net features more effectively. Because the proposed algorithm learns to estimate metric depths by exploiting gradient maps and relative depths, additional datasets containing only relative depths can be used to improve the metric depth estimation performance further. Experimental results show that the proposed algorithm is competitive with state-of-the-art metric depth estimators, even when it is trained with a smaller metric depth dataset.

This paper has the following contributions:
\begin{itemize}
    \itemsep0em
    \item We propose a novel monocular depth estimator, which decomposes a metric depth map into a normalized depth map and relative depth features and then exploits those relative features to improve the metric depth estimation performance.

    \item The proposed algorithm can be adapted to a new camera efficiently since it can be trained with a small metric depth dataset together with camera-independent relative depth datasets.

    \item The proposed algorithm provides competitive performance to conventional state-of-the-art metric depth estimators and can improve the performance further through joint training using multiple datasets.
\end{itemize}
\section{Related Work}
\label{sec:related_work}

\subsection{Monocular Metric Depth Estimation}

The objective of monocular metric depth estimation is to predict pixel-wise absolute distances of a scene from a camera using a single image. Since different 3D scenes can be projected onto the same 2D image, monocular depth estimation is ill-posed. Nevertheless, active research has been conducted due to its practical importance. To infer depths, early approaches made prior assumptions on scenes, \eg\ box blocks \cite{gupta2010eccv}, planar regions \cite{saxena2008make3d}, or particular layout of objects \cite{gupta2010nips}. However, they may provide implausible results, especially in regions with ambiguous colors or small objects.

With recent advances in deep learning, CNN techniques for monocular depth estimation have been developed, yielding excellent performance. Many attempts have been made to find better network architecture \cite{eigen2014depth, laina2016deeper, xu2017multi, heo2018monocular, chen2019structure} or to design more effective loss functions \cite{eigen2015predicting, chen2016single, laina2016deeper, hu2019revisiting}. It has been also demonstrated that the depth estimation performance can be improved by predicting quantized depths through ordinal regression \cite{fu2018deep}, by employing Fourier domain analysis \cite{lee2018single}, by enforcing geometric constraints of virtual normals \cite{yin2019enforcing}, or by reweighting multiple loss functions \cite{lee2020multi}. Recently, the vision transformer \cite{dosovitskiy2020image} was employed for monocular depth estimation \cite{bhat2021adabins}, improving the performance significantly.

\subsection{Relative Depth Estimation}
The objective of relative depth estimation is to learn the pairwise depth order \cite{zoran2015learning} or the rank of pixels \cite{chen2016single, xian2020structure} in an image. Recently, listwise ranking, instead of pairwise ranking, was considered for relative depth estimation \cite{lienen2021monocular}. Also, scale-invariant loss \cite{eigen2014depth} and its variants \cite{li2018megadepth, li2019learning, wang2019web, ranftl2020} have been used to alleviate the scale ambiguity of depths, thereby improving the performance of relative depth estimation.

Unlike metric depths, relative depth information --- or depth order information --- is invariant to camera parameters. Therefore, even though a training set is composed of images captured by different cameras, it does not affect the performance of relative depth estimation adversely. Therefore, heterogeneous training data, such as disparity maps from stereo image pairs \cite{wang2019web, xian2018monocular, xian2020structure} or videos \cite{ranftl2020}, structure-from-motion reconstruction \cite{li2018megadepth, li2019learning}, and ordinal labels \cite{chen2016single}, have been used to train relative depth estimators.

\subsection{Relative vs.\ Metric Depths}
A metric depth map contains relative depth information, whereas relative depth information is not sufficient for reconstructing a metric depth map. However, relative-to-metric depth conversion has been attempted by fitting relative depths to metric depths \cite{ranftl2020, lienen2021monocular} or by fine-tuning a relative depth estimator for metric depth estimation \cite{ranftl2021vision}.

On the other hand, relative and metric depths can be jointly estimated to exploit their correlation and to eventually improve the performance of metric depth estimation. To this end, ordinal labels are used with a ranking loss in \cite{chen2016single}. Also, in \cite{lee2019monocular}, relative and metric depth maps at various scales are first estimated and then optimally combined to yield a final metric depth map.

The proposed algorithm also estimates relative depth information, in addition to metric depths, to improve the performance of metric depth estimation. However, differently from \cite{chen2016single, lee2019monocular}, the proposed algorithm better exploits the correlation between relative and metric depths by decomposing a metric depth map. Furthermore, the proposed algorithm can provide promising results even with a small metric depth dataset by exploiting a relative depth dataset additionally.

\begin{figure}[!t]
  \centering
   \includegraphics[width=1\linewidth]{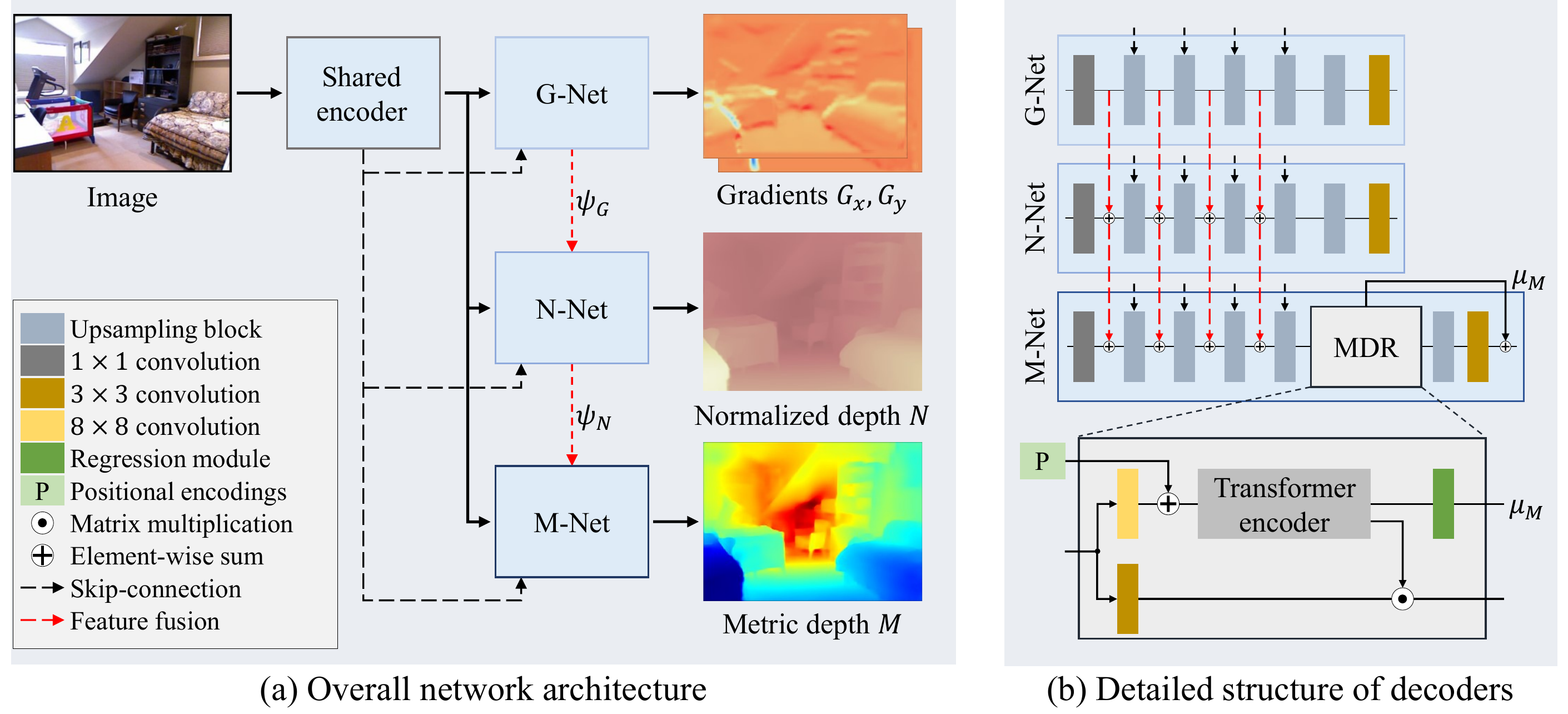}
   \caption{ (a) Overall network architecture of the proposed algorithm and (b) detailed structure of decoders. The proposed algorithm consists of a shared encoder and three decoders: G-Net, N-Net, and M-Net. G-Net predicts horizontal and vertical gradient maps, while N-Net and M-Net estimate normalized and metric depth maps, respectively. Note that G-Net features are fed into N-Net, and N-Net features are fed into M-Net.}
   \label{fig:overview}
\end{figure}

\section{Proposed Algorithm}
\label{sec:proposd_algorithm}

Fig.~\ref{fig:overview} is an overview of the proposed algorithm, which consists of a shared encoder and three decoders --- G-Net, N-Net, and M-Net. The shared encoder extracts common features that are fed into the three decoders. Then, G-Net predicts horizontal and vertical gradients of depths, while N-Net and M-Net estimate a normalized depth map and a metric depth map, respectively. Note that features extracted by G-Net are fed into N-Net to convey edge information, and those by N-Net are, in turn, fed into M-Net to provide relative depth features. Finally, via the MDR block, M-Net exploits the relative depth features to estimate a metric depth map more accurately.

\subsection{Metric Depth Decomposition}
\label{sec:Metric_Depth_Decomposition}

Given an RGB image $I \in \mathbb{R}^{h \times w \times 3}$, the objective is to estimate a metric depth map $M \in \mathbb{R}^{h \times w}$. However, this is ill-posed because different scenes with different metric depths can be projected onto the same image. Moreover, scale features of depths are hard to estimate from the color information only since they also depend on the camera parameters. To address this issue, we decompose a metric depth map $M$ into a normalized depth map $N$ and scale parameters. The normalized depth map $N$ contains relative depth information, so it is less sensitive to scale variations or camera parameters than the metric depth map $M$ is.

There are several design choices for normalizing a metric depth map, including min-max normalization or ranking-based normalization \cite{han2011data}. However, the min-max normalization is sensitive to outliers, and the ranking-based normalization is unreliable in areas with homogeneous depths, such as walls and floors. Instead, we normalize a metric depth map using the $z$-score normalization. Given a metric depth map $M$, we obtain the normalized depth map $N$ by
\begin{equation}
    N = \frac{M - \mu_{M} U}{\sigma_{M}}
    \label{eq:M2N}
\end{equation}
where $\mu_M$ and $\sigma_M$, respectively, denote the mean and standard deviation of metric depths in $M$. Also, $U$ is the unit matrix whose all elements are 1.

N-Net, denoted by $f_N$, estimates this normalized depth map, and its estimate is denoted by $\hat{N}$. When the scale parameters $\mu_M$ and $\sigma_M$ are known, the metric depth map $M$ can be reconstructed by
\begin{equation}
    \hat{M}_{\textrm{direct}} = \sigma_{M} \hat{N} + \mu_{M} U.
    \label{eq:N2M}
\end{equation}
In practice, $\mu_M$ and $\sigma_M$ are unknown. Conventional methods in \cite{ranftl2020, lienen2021monocular, ranftl2021vision} obtain fixed $\sigma_{M}$ and $\mu_{M}$ for all images based on the least-squares criterion. In such a case, the accuracy of $\hat{M}_{\textrm{direct}}$ in \eqref{eq:N2M} greatly depends on the accuracy of $\sigma_{M}$ and $\mu_{M}$. In this work, instead of the direct conversion in \eqref{eq:N2M}, we estimate the metric depth map by employing the features $\psi_N(I)$, which are extracted by the proposed M-Net, $f_N$, during the estimation of $\hat{N}$. In other words, the proposed M-Net, $f_M$, estimates the metric depth map by
\begin{equation}
\hat{M} = f_M(I, \psi_N(I)).
    \label{eq:M_Net}
\end{equation}

For metric depth estimation, structural data (\eg\ surface normals or segmentation maps) have been adopted as additional cues \cite{eigen2015predicting, qi2018geonet, liu2019end, lee2021learning}, or relative depth features have been used indirectly via loss functions (\eg\ pairwise ranking loss \cite{chen2016single} or scale-invariant loss \cite{eigen2014depth, lee2019big, bhat2021adabins}). In contrast, we convert a metric depth map to a normalized depth map. Then, the proposed N-Net  estimates the normalized depth map to extract the features $\psi_N$, containing relative depth information. Then, the proposed M-Net uses $\psi_N$ for effective metric depth estimation.

Similarly, we further decompose the normalized depth map $N$ into more elementary data: horizontal and vertical gradients. The horizontal gradient map $G_x$ is given by
\begin{equation}
    G_{x} = \nabla_x N
\end{equation}
where $\nabla_x$ is the partial derivative operator computing the differences between horizontally adjacent pixels. The vertical gradient map $G_y$ is obtained similarly. The proposed G-Net is trained to estimate these gradient maps $G_x$ and $G_y$. Hence, G-Net learns edge information in a scene, and its features $\psi_G$ are useful for inferring the normalized depth map. Therefore, similar to \eqref{eq:M_Net}, N-Net estimates the normalized depth map via
\begin{equation}
    \hat{N} = f_N(I, \psi_G(I))
\label{eq:G_Net}
\end{equation}
using the gradient features $\psi_G(I)$.

\subsection{Network Architecture}
For the shared encoder in Fig.~\ref{fig:overview}, we adopt EfficientNet-B5 \cite{tan2019efficientnet} as the backbone network. G-Net and N-Net have an identical structure, consisting of five upsampling blocks. However, G-Net outputs two channels for two gradient maps $G_x$ and $G_y$, while N-Net yields a single channel for a normalized depth map $N$. M-Net also has a similar structure, except for the MDR block, which will be detailed in Section~\ref{sec:MDR_Block}. MDR predicts the mean $\mu_M$ of $M$ separately, which is added back at the end of M-Net.

The encoder features are fed into the three decoders via skip-connections \cite{he2016deep}, as shown in Fig.~\ref{fig:overview}(a). To differentiate the encoder features for the different decoders, we apply $1 \times 1$ convolution to the encoder features before feeding them to each decoder. Also, we apply the channel attention \cite{hu2018squeeze} before each skip-connection to each decoder.

We transfer features unidirectionally from G-Net to N-Net and also from N-Net to M-Net to exploit low-level features for the estimation of high-level data. To this end, we fuse features through element-wise addition before each of the first four upsampling blocks in N-Net and M-Net, as shown in Fig.~\ref{fig:overview}(b). Specifically, let $\psi_G^\mathrm{out}$ and $\psi_N^\mathrm{out}$ denote the output features of G-Net and N-Net, respectively. Then, the input feature $\psi^\mathrm{in}_N$ to the next layer of N-Net is given by
\begin{equation}
    \psi^\mathrm{in}_N = \omega_G \psi_G^\mathrm{out} + \omega_N \psi_N^\mathrm{out}
    \label{eq:feature_fusion}
\end{equation}
where $\omega_G$ and $\omega_N$ are pre-defined weights for $\psi_G^\mathrm{out}$ and $\psi_N^\mathrm{out}$ to control the relative contributions of the two features. For M-Net, the features from N-Net are added similarly. In order to fuse features, we use addition, instead of multiplication or concatenation, for computational efficiency.

\begin{figure}[!t]
    \centering
    \includegraphics[width=\linewidth]{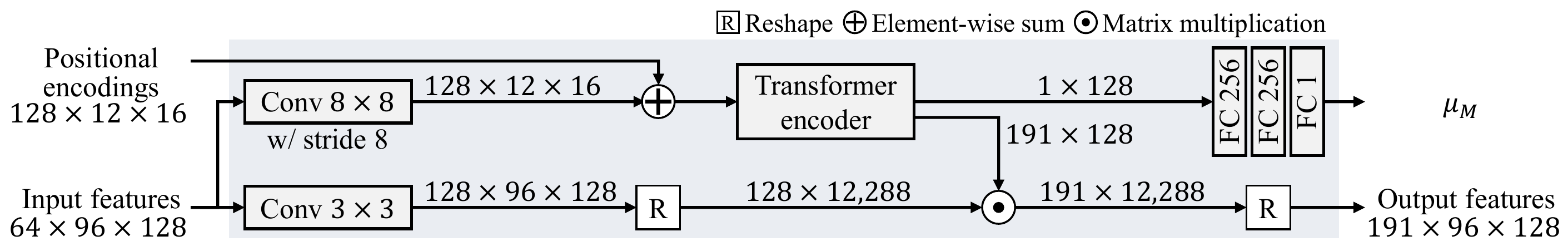}
    \caption{Detailed structure of the MDR block.}
    \label{fig:MDR_block}
\end{figure}

\subsection{MDR Block}
\label{sec:MDR_Block}
We design the MDR block to utilize the features $\psi_N$ of N-Net more effectively for the metric depth estimation in M-Net. Fig.~\ref{fig:MDR_block} shows the structure of the MDR block, which applies patchwise attention to an input feature map and estimates the mean $\mu_M$ of $M$ separately using the transformer encoder \cite{dosovitskiy2020image}. Note that the transformer architecture enables us to use one of the patchwise-attended feature vectors to regress~$\mu_M$.

Specifically, MDR first processes input features using an $8 \times 8$ convolution layer with a stride of 8 and a $3 \times 3$ convolution layer, respectively. The patchwise output of the $8 \times 8$ convolution is added to the positional encodings and then input to the transformer encoder \cite{dosovitskiy2020image}. The positional encodings are learnable parameters, randomly initialized at training. Then, the transformer encoder generates 192 patchwise-attended feature vectors of 128 dimensions. We adopt the mini-ViT architecture \cite{bhat2021adabins} for the transformer encoder. The first vector is fed to the regression module, composed of three fully-connected layers, to yield $\mu_M$. The rest 191 vectors form a matrix, which is multiplied with the output of the $3 \times 3$ convolution layer to generate $191 \times 96 \times 128$ output features through reshaping. Finally, those output features are fed to the next upsampling block of M-Net.
Also, the estimated $\mu_M$ is added back at the end of M-Net in Fig.~\ref{fig:overview}(b), which makes the remaining parts of M-Net focus on the estimation of the mean-removed depth map $M-\mu_M U$ by exploiting the N-Net features $\psi_N$.

\subsection{Loss Functions}
Let us describe the loss functions for training the three decoders. For G-Net, we use the $\ell_1$ loss
\begin{equation}
    \mathcal{L}_G = \frac{1}{T}\big(\Vert \hat{G}_x - G_x\Vert_1 + \Vert \hat{G}_y - G_y\Vert_1\big)
\end{equation}
where $\hat{G}_x$ and $\hat{G}_y$ are predictions of the ground-truth gradient maps $G_x$ and $G_y$, respectively. Also, $T$ denotes the number of valid pixels in the ground-truth.

For N-Net, we use two loss terms: the $\ell_1$ loss and the gradient loss. The $\ell_1$ loss is defined as
\begin{equation}
    \mathcal{L}_N = \frac{1}{T}\Vert \hat{N} - N\Vert_1
\end{equation}
where $\hat{N}$ and $N$ are predicted and ground-truth normalized depth maps. Note that scale-invariant terms are often adopted to train monocular depth estimators \cite{lee2019big, bhat2021adabins, ranftl2021vision}. However, we do not use such scale-invariant losses since normalized depth maps are already scale-invariant. Next, the gradient losses \cite{li2018megadepth, xian2020structure, hu2019revisiting} for $N$ in the horizontal direction are defined as
\begin{equation}
    \mathcal{L}_{Nx} = \frac{1}{T \cdot s^2}\Vert \nabla_x\hat{N}_s - \nabla_x N_s\Vert_1
\end{equation}
where $\hat{N}_s$ and $N_s$ are the bilinearly scaled $\hat{N}$ and $N$ with a scaling factor $s$. We compute the gradient losses at various scales, as in \cite{li2018megadepth, xian2020structure}, by setting $s$ to 0.5, 0.25, and 0.125. The losses $L_{Ny}$ in the vertical direction are also used.

Similarly, for M-Net, we use the loss terms $\mathcal{L}_M$, $\mathcal{L}_{Mx}$, and $\mathcal{L}_{My}$. In addition, we use two more loss terms. First, $\mathcal{L}_{\mu_M}$ is defined to train the MDR block, which is given by
\begin{equation}
    \mathcal{L}_{\mu_M} = \Vert \mu(\hat{M}) - \mu_M\Vert_1
\end{equation}
where $\mu(\hat{M})$ denotes the mean of depth values in $\hat{M}$. Second, we define the logarithmic $\ell_1$ loss,
\begin{equation}
    \mathcal{L}_{\log M} = \frac{1}{T}\Vert \log{\hat{M}} - \log{M}\Vert_1.
    \label{eq:loss_logarithmic}
\end{equation}

In this work, we adopt inverse depth representation of metric depths to match the depth order with a relative depth dataset \cite{xian2020structure}. In this case, theoretically, a metric depth can have a value in the range $[0, \infty)$. Thus, when a metric depth is near zero, its inverted value becomes too large, which interferes with training. We overcome this problem through a simple modification. Given an original metric depth $m_o$, its inverted metric depth $m$ is defined as
\begin{equation}
    m = 1 / (m_o + 1).
    \label{eq:inverted_depth}
\end{equation}
In this way, inverted metric depth values are within the range of $(0, 1]$ and also are more evenly distributed.

However, using the $\ell_1$ loss $\mathcal{L}_M$ on inverse depths has a disadvantage in learning distant depths. Suppose that $\hat{\chi}$ and $\chi$ are predicted and ground-truth metric depth values for a pixel, respectively. Then, the $\ell_1$ error $E$ is given by
\begin{equation}
    E =  \bigg| \frac{1}{\hat{\chi}} - \frac{1}{\chi}\bigg|.
    \label{sec:loss_L1_inv_example}
\end{equation}
As $\chi$ gets larger, $E$ becomes smaller for the same $| \hat{\chi} - \chi |$. This means that the network is trained less effectively for distant regions. This problem is alleviated by employing $\mathcal{L}_{\log M}$ in \eqref{eq:loss_logarithmic}.

\section{Experimental Results}
\label{sec:experimental_results}

\subsection{Datasets}
We use four depth datasets: one for relative depths~\cite{xian2020structure} and three for metric depths~\cite{silberman2012indoor,kim2018deep,song2015sun}. When relative depth data are used in training, losses are generated from the loss terms for N-Net and G-Net only because the loss terms for M-Net cannot be computed.

{\noindent \bf HR-WSI}~\cite{xian2020structure}: It consists of 20,378 training and 400 test images. The ground-truth disparity maps are generated by FlowNet 2.0~\cite{ilg2017flownet}. We use only the training data of HR-WSI. We normalize the disparity maps by \eqref{eq:M2N} and regard them as normalized depth maps.

{\noindent \bf NYUv2}~\cite{silberman2012indoor}: It contains 120K video frames for training and 654 frames for test, together with the depth maps captured by a Kinect v1 camera. We use the NYUv2 dataset for both training and evaluation. We construct three training datasets of 51K, 17K, and 795 sizes. Specifically, we extract the 51K and 17K images by sampling video frames uniformly. For the 795 images, we use the official training split of NYUv2. We fill in missing depths using the colorization scheme~\cite{levin2004colorization}, as in~\cite{silberman2012indoor}.

{\noindent \bf DIML-Indoor}~\cite{kim2018deep}: It consists of 1,609 training images and 503 test images, captured by a Kinect v2 camera.

{\noindent \bf SUN RGB-D}~\cite{song2015sun}: It consists 5,285 training images and 5,050 test images, obtained by four different cameras: Kinect v1, Kinect v2, RealSense, and Xtion.

\subsection{Evaluation Metrics}
{\noindent \bf Metric depths}: We adopt the four evaluation metrics in~\cite{eigen2014depth}, listed below. Here, $M_i$ and $\hat{M}_i$ denote the ground-truth and predicted depths of pixel $i$, respectively. $|\cdot|$ denotes the number of valid pixels in a depth map. For the NYUv2 dataset, we adopt the center crop protocol~\cite{eigen2014depth}.
{\footnotesize
\begin{align}
\textstyle
\textrm{RMSE}: & \ \ \frac{1}{|M|}\big(\sum_{i}(\hat{M}_i-M_i)^2\big)^{0.5} \nonumber \\
\textrm{REL}:  & \ \ \frac{1}{|M|}\sum_{i} \vert \hat{M}_i-M_i \vert /M_i \nonumber \\
\log10:        & \ \ \frac{1}{|M|}\sum_{i} \vert \log_{10}(\hat{M}_i)-\log_{10}(M_i) \vert \nonumber \\
\delta_k:      & \ \ \textrm{\% of $M_i$ that satisfies} \ \max\!\left(\frac{\hat{M}_i}{M_i},\frac{M_i}{\hat{M}_i}\right)\!<1.25^k, \ \ k \in \{1, 2, 3\} \nonumber
\end{align}}

{\noindent \bf Relative depths:} we use two metrics for relative depths. First, WHDR (weighted human disagreement rate)~\cite{chen2016single, ranftl2020, xian2020structure} measures the ordinal consistency between point pairs. We follow the evaluation protocol of~\cite{xian2020structure} to randomly sample 50,000 pairs in each depth map. However, WHDR is an unstable protocol, under which the performance fluctuates with each measurement. We hence use Kendall's $\tau$~\cite{kendall1938new} additionally, which considers the ordering relations of all pixel pairs. Given a ground-truth normalized depth map $D$ and its prediction $\hat{D}$, Kendall's $\tau$ is defined as
\begin{equation}
    \tau{(\hat{D}, D)} = \frac{\alpha(\hat{D}, D) - \beta(\hat{D}, D)}{\binom{|D|}{2}}
\end{equation}
where $\alpha(\hat{D}, D)$ and $\beta(\hat{D}, D)$ are the numbers of concordant pairs and discordant pairs between $D$ and $\hat{D}$, respectively. Note that Kendall's $\tau$ can measure the quality of a metric depth map, as well as that of a relative one.

\begin{table}[!t]
    \scriptsize
    \addtolength{\tabcolsep}{0.2pt}
    \renewcommand{\arraystretch}{1.1}
    \caption{Comparison of depth estimation results on the NYUv2 dataset. `\#' is the number of training images, and $\dagger$ means that additional data is used for training. The best results are \textbf{boldfaced}. Lower RMSE, REL, and $\log10$ indicate better results, while higher $\delta_k$ values are better ones.}

    \centering
    \begin{tabular}{l|c|r|cccccc}
    \toprule
    & \# & \multicolumn{1}{c|}{Encoder backbone} & \ RMSE & REL & $\log10$ & \ $\delta_1$ \ & \ $\delta_2$ \ &  \ $\delta_3$ \ \\
    \midrule
    \multicolumn{1}{l|}{Eigen~\etal~\cite{eigen2014depth}} & \ 120K \ & - \hspace{1cm} & 0.641 & 0.158 & - & 0.769 & 0.950 & 0.988\\
    \multicolumn{1}{l|}{Laina~\etal~\cite{laina2016deeper}} & 12K & ResNet-50~\cite{he2016deep} \ & 0.573 & 0.127 & 0.055 & 0.811 & 0.953 & 0.988\\
    \multicolumn{1}{l|}{Hao~\etal~\cite{hao2018detail}} & 13K & ResNet-101~\cite{he2016deep} \ & 0.555 & 0.127 & 0.053 & 0.841 & 0.966 & 0.991\\
    \multicolumn{1}{l|}{Fu~\etal~\cite{fu2018deep}} & 120K & ResNet-101~\cite{he2016deep} \ & 0.509 & 0.115 & 0.051 & 0.828 & 0.965 & 0.992\\
    \multicolumn{1}{l|}{Hu~\etal~\cite{hu2019revisiting}} & 50K & SENet-154~\cite{hu2018squeeze} \ & 0.530 & 0.115 & 0.050 & 0.866 & 0.975 & 0.993\\
    \multicolumn{1}{l|}{Chen~\etal~\cite{chen2019structure}} & 50K & SENet-154~\cite{hu2018squeeze} \ & 0.514 & 0.111 & 0.048 & 0.878 & 0.977 & 0.994\\
    \multicolumn{1}{l|}{Yin~\etal~\cite{yin2019enforcing}} & 29K & ResNeXt-101~\cite{xie2017aggregated} \ & 0.416 & 0.108 & 0.048 & 0.875 & 0.976 & 0.994\\
    \multicolumn{1}{l|}{Lee~\etal~\cite{lee2019big}} & 24K & DenseNet-161~\cite{huang2017densely} \ & 0.392 & 0.110 & 0.047 & 0.885 & 0.978 & 0.994\\
    \multicolumn{1}{l|}{Hyunh~\etal~\cite{huynh2020guiding}} & 50K & DRN-D-22~\cite{yu2017dilated} \ & 0.412 & 0.108 & - & 0.882 & 0.980 & 0.996\\
    \multicolumn{1}{l|}{Lee and Kim~\cite{lee2020multi}} & 58K & PNASNet-5~\cite{liu2018progressive} \ & 0.430 & 0.119 & 0.050 & 0.870 & 0.974 & 0.993\\
    \multicolumn{1}{l|}{Bhat~\etal~\cite{bhat2021adabins}} & 50K & \ EfficientNet-B5~\cite{tan2019efficientnet} \ & 0.364 & 0.103 & 0.044 & 0.903 & 0.984 & \textbf{0.997}\\
    \multicolumn{1}{l|}{Proposed} & 51K & EfficientNet-B5~\cite{tan2019efficientnet} \ & \textbf{0.362} & \textbf{0.100} & \textbf{0.043} & \textbf{0.907} & \textbf{0.986} & \textbf{0.997}\\
    \midrule\midrule
    \multicolumn{1}{l|}{Wang~\etal~\cite{wang2015towards}$^\dagger$} & 200K & - \hspace{1cm} & 0.745 & 0.220 & 0.094 & 0.605 & 0.890 & 0.970\\
    \multicolumn{1}{l|}{Ramam. and Lepetit~\cite{ramamonjisoa2019sharpnet}$^\dagger$} & 400K & ResNet-50~\cite{he2016deep} \ & 0.502 & 0.139 & 0.047 & 0.836 & 0.966 & 0.993\\
    \multicolumn{1}{l|}{Ranftl~\etal~\cite{ranftl2021vision}$^\dagger$} & 1.4M & ViT-Hybrid{\hskip 0.5em}~\cite{dosovitskiy2020image} \ & 0.357 & 0.110 & 0.045 & 0.904 & \textbf{0.988} & \textbf{0.998}\\
    \multicolumn{1}{l|}{Proposed$^\dagger$} & 71K & EfficientNet-B5~\cite{tan2019efficientnet} \ & \textbf{0.355} & \textbf{0.098} & \textbf{0.042} & \textbf{0.913} & 0.987 & \textbf{0.998}\\
    \bottomrule
    \end{tabular}
    \label{tb:performance_NYU}
\end{table}

\begin{figure}[!t]
  \centering
   \includegraphics[width=0.98\linewidth]{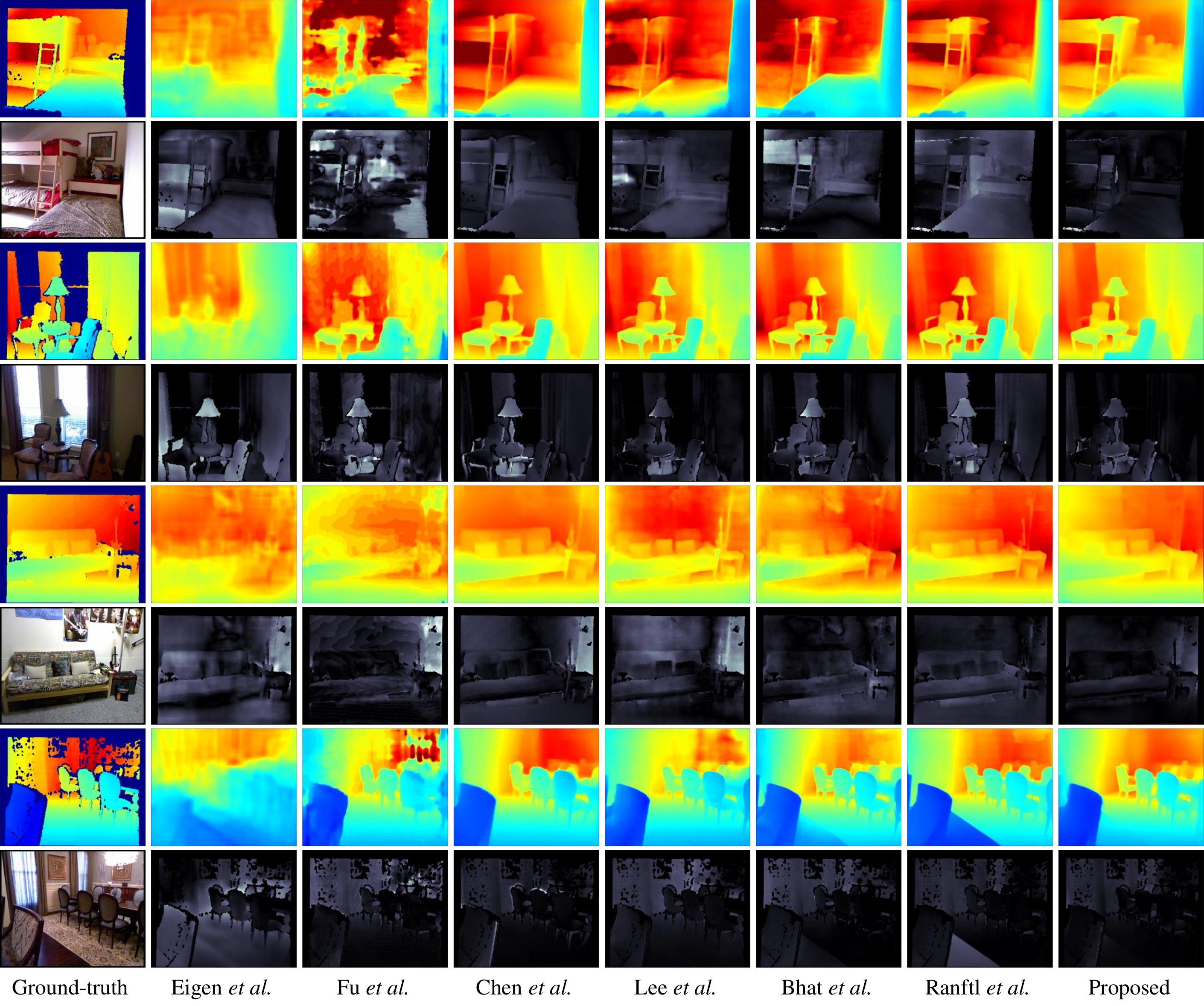}
   \caption{
   Qualitative comparison of the proposed algorithm with the conventional algorithms. For each depth map, the error map is also provided, in which brighter pixels correspond to larger errors.
   }
   \label{fig:qualitative}
\end{figure}

\subsection{Implementation Details}
{\noindent \bf Network architecture:} We employ EfficientNet-B5 \cite{tan2019efficientnet} as the encoder backbone. The encoder takes an $512 \times 384$ RGB image and generates a $16 \times 12$ feature with 2,048 channels. The output feature is used as the input to the three decoders. G-Net and N-Net consist of 5 upsampling blocks, each of which is composed of a bilinear interpolation layer and two $3\times 3$ convolution layers with the ReLU activation. Also, in addition to the 5 upsampling blocks, M-Net includes the MDR block, located between the fourth and fifth upsampling blocks. For feature fusion in \eqref{eq:feature_fusion}, $\omega_G = \omega_N = 1$.

{\noindent \bf Training:} We train the proposed algorithm in two phases. First, we train the network, after removing M-Net, for 20 epochs with an initial learning rate of $10^{-4}$. The learning rate is decreased by a factor of 0.1 at every fifth epoch. Second, we train the entire network, including all three decoders, jointly for 15 epochs with an initial learning rate of $10^{-4}$, which is decreased by a factor of 0.1 at every third epoch. We use the Adam optimizer~\cite{kingma2014adam} with a weight decay of $10^{-4}$. If a relative depth is used in the second phase, losses are calculated from the loss terms for N-Net and G-Net only.

\subsection{Performance Comparison}
Table~\ref{tb:performance_NYU} compares the proposed algorithm with conventional ones on NYUv2 dataset. Some of the conventional algorithms use only NYUv2 training data \cite{eigen2014depth,laina2016deeper,hao2018detail,fu2018deep,hu2018squeeze,chen2019structure,yin2019enforcing,lee2019big,huynh2020guiding,lee2020multi,bhat2021adabins}, while the others use extra data~\cite{wang2015towards,ranftl2021vision,ramamonjisoa2019sharpnet}. For fair comparisons, we train the proposed algorithm in both ways: `Proposed' uses NYUv2 only, while `Proposed$^\dagger$' uses both HR-WSI and NYUv2. The following observations can be made from Table~\ref{tb:performance_NYU}.
\begin{itemize}
\itemsep0mm
\item `Proposed' outperforms all conventional algorithms in all metrics with no exception.
For example, `Proposed' provides a REL score of 0.100, which is 0.003 better than that of the second-best algorithm, Bhat~\etal \cite{bhat2021adabins}. Note that both algorithms use the same encoder backbone of EfficientNet-B5 \cite{tan2019efficientnet}.
\item `Proposed$^\dagger$' provides the best results in five out of six metrics. For $\delta_2$, the proposed algorithm yields the second-best score after Ranftl \etal\ \cite{ranftl2021vision}. It is worth pointing out that Ranftl \etal\ uses about 20 times more training data than the proposed algorithm does.
\end{itemize}
Fig.~\ref{fig:qualitative} compares the proposed algorithm with the conventional algorithms \cite{eigen2014depth, fu2018deep, chen2019structure, lee2019big, bhat2021adabins, ranftl2021vision} qualitatively. We see that the proposed algorithm estimates the depth maps more faithfully with smaller errors.

\begin{table}[!t]
    \scriptsize
    \addtolength{\tabcolsep}{2.7pt}
    \renewcommand{\arraystretch}{1.1}
    \caption{Comparison of depth estimation results on various datasets. `\#' is the size of metric depth dataset, and $\dagger$ means that 20K HR-WSI data are additionally used for training. A lower Kendall's $\tau$ indicates a better result.}
    \centering
    \begin{tabular}{c@{\hskip 0.9em}c@{\hskip 0.9em}c@{\hskip 0.9em}c@{\hskip 0.9em}c@{\hskip 0.9em}c@{\hskip 0.9em}c@{\hskip 0.9em}|ccccccc}
    \toprule
    \hspace{1.5cm} & \multicolumn{4}{c}{\#} & \multicolumn{2}{c}{Setting} & RMSE & REL & $\log10$ & $\delta_1$ & $\delta_2$ & $\delta_3$ & Kendall's $\tau$\\
    \midrule
    \multicolumn{1}{c}{\multirow{10}{*}{NYUv2}}
    & \multicolumn{4}{c}{\multirow{3}{*}{795}} & \multicolumn{2}{|l|}{Baseline} & 0.487 & 0.147 & 0.061 & 0.809 & 0.963 & 0.991 & 0.738\\
    & \multicolumn{4}{c}{} & \multicolumn{2}{|l|}{Proposed} & 0.468 & 0.142 & 0.059 & 0.824 & 0.969 & 0.992 & 0.762\\
    & \multicolumn{4}{c}{} & \multicolumn{2}{|l|}{Proposed$^\dagger$} & \textbf{0.417} & \textbf{0.122} & \textbf{0.052} & \textbf{0.868} & \textbf{0.977} & \textbf{0.995} & \textbf{0.800}\\ \cmidrule(lr){2-14}
    & \multicolumn{4}{c}{\multirow{3}{*}{17K}} & \multicolumn{2}{|l|}{Baseline} & 0.400 & 0.113 & 0.048 & 0.880 & 0.981 & 0.996 & 0.803\\
    & \multicolumn{4}{c}{} & \multicolumn{2}{|l|}{Proposed} & 0.370 & 0.103 & 0.045  & 0.903 & 0.986 & \textbf{0.997}& 0.829\\
    & \multicolumn{4}{c}{} & \multicolumn{2}{|l|}{Proposed$^\dagger$} & \textbf{0.362} & \textbf{0.100} & \textbf{0.043} & \textbf{0.909} & \textbf{0.987} & \textbf{0.997} & \textbf{0.835}\\ \cmidrule(lr){2-14}
    & \multicolumn{4}{c}{\multirow{3}{*}{51K}} & \multicolumn{2}{|l|}{Baseline} & 0.386 & 0.109 & 0.047 & 0.888 & 0.980 & 0.995 & 0.813\\
    & \multicolumn{4}{c}{} & \multicolumn{2}{|l|}{Proposed} & 0.362 & 0.100 & 0.043 & 0.907 & 0.986 & 0.997 & 0.837\\
    & \multicolumn{4}{c}{} & \multicolumn{2}{|l|}{Proposed$^\dagger$} & \textbf{0.355} & \textbf{0.098} & \textbf{0.042} & \textbf{0.913} & \textbf{0.987} & \textbf{0.998} & \textbf{0.840}\\
    \midrule
    \multicolumn{1}{c}{\multirow{3}{*}{DIML-Indoor}}
    & \multicolumn{4}{c}{\multirow{3}{*}{1.6K}} & \multicolumn{2}{|l|}{Baseline} & 0.589 & 0.247 & 0.099 & 0.701 & 0.879 & 0.968 & 0.492\\
    & \multicolumn{4}{c}{} & \multicolumn{2}{|l|}{Proposed} & 0.537 & 0.180 & 0.075 & 0.719 & 0.943 & 0.986 & 0.696\\
    & \multicolumn{4}{c}{} & \multicolumn{2}{|l|}{Proposed$^\dagger$} & \textbf{0.517} & \textbf{0.171} & \textbf{0.072} & \textbf{0.742} & \textbf{0.949} & \textbf{0.989} & \textbf{0.742}\\
    \midrule
    \multicolumn{1}{c}{\multirow{3}{*}{SUN RGB-D}}
    & \multicolumn{4}{c}{\multirow{3}{*}{5.3K}} & \multicolumn{2}{|l|}{Baseline} & 0.306 & 0.132 & 0.055 & 0.847 & 0.971 & 0.992 & 0.761\\
    & \multicolumn{4}{c}{} & \multicolumn{2}{|l|}{Proposed} & 0.303 & 0.129 & 0.055 & 0.850 & \textbf{0.973} & \textbf{0.993} & 0.776\\
    & \multicolumn{4}{c}{} & \multicolumn{2}{|l|}{Proposed$^\dagger$} & \textbf{0.301} & \textbf{0.127} & \textbf{0.054} & \textbf{0.853} & \textbf{0.973} & 0.992 & \textbf{0.784}\\
    \bottomrule
    \end{tabular}
    \label{tb:dataset_size_nyu}
\end{table}

\begin{figure}[!t]
  \centering
   \includegraphics[width=\linewidth]{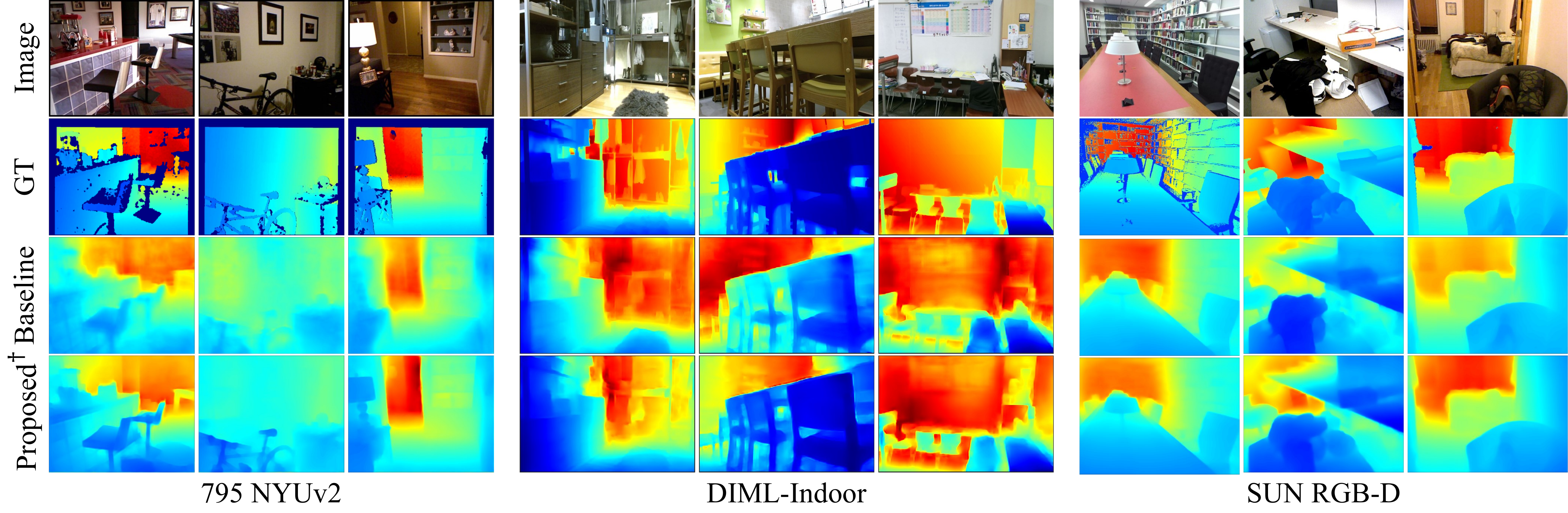}
   \caption{Qualitative comparison of the proposed algorithm with the baseline using the NYUv2 (795), DIML-Indoor, and SUN RGB-D datasets.}
   \label{fig:qualitative_795}
\end{figure}

\subsection{Various Datasets}
Table~\ref{tb:dataset_size_nyu} verifies the effectiveness of the proposed algorithm on various datasets.
The first two columns in Table~\ref{tb:dataset_size_nyu} indicate the metric depth dataset and its size. `Baseline' is a bare encoder-decoder for monocular depth estimation. Specifically, we remove G-Net and N-Net, as well as the MDR block in M-Net, from the proposed algorithm in Fig.~\ref{fig:overview} to construct `Baseline.' For its training, only three loss terms $\mathcal{L}_M$, $\mathcal{L}_{Mx}$, and $\mathcal{L}_{My}$ are used. `Proposed' means the proposed algorithm without employing the 20K HR-WSI training data, while `Proposed$^\dagger$' means using the HR-WSI data additionally. The following observations can be made from Table~\ref{tb:dataset_size_nyu}.
\begin{itemize}
\itemsep0mm
\item By comparing `Proposed' with `Baseline,' we see that G-Net and N-Net help M-Net improve the performance of metric depth estimation by transferring edge information and relative depth information. Also, `Proposed$^\dagger$' meaningfully outperforms `Proposed' by leveraging relative depth training data in HR-WSI, which contain no metric depth labels.
\item Even when only the 795 NYUv2 images are used, the proposed algorithm provides acceptable results. For example, the RMSE score of 0.417 is similar to that of the Hyunh \etal's estimator \cite{huynh2020guiding} in Table~\ref{tb:performance_NYU}, which uses 50K metric depth map data. In contrast, the proposed algorithm uses the 795 metric depth maps only.
\item The proposed algorithm also exhibits similar trends in the DIML-Indoor and SUN RGB-D datasets, which are collected using different cameras: the proposed algorithm can be trained effectively even with a small number of metric depth images. This is advantageous in practical applications in which an algorithm should be adapted for various cameras.
\end{itemize}
Fig.~\ref{fig:qualitative_795} compares `Baseline' and `Proposed$^\dagger$' qualitatively using the 795 NYUv2, DIML-Indoor, and SUN RGB-D datasets. For all datasets, `Proposed$^\dagger$' provides more accurate and more detailed depth maps, especially around chairs, tables, and desks, than `Baseline' does.

\subsection{Analysis}
\begin{table}[!t]
    \scriptsize
    \addtolength{\tabcolsep}{1.2pt}
    \renewcommand{\arraystretch}{1.1}
    \caption{Ablation studies of the proposed algorithm using the NYUv2 (17K) dataset.}
    \centering
    \begin{tabular}{cccccc|cccccccc}
    \toprule
    M & N & G & MDR$^*$ & MDR & $\dagger$ & RMSE & REL & $\log10$ & $\delta_1$ & $\delta_2$ & $\delta_3$ & Kendall's $\tau$ & WHDR(\%)\\
    \midrule
    \checkmark & - & - & - & - & - & 0.400 & 0.113 & 0.048 & 0.880 & 0.981 & 0.996 & 0.803 & 14.95\\
    \checkmark & \checkmark & - & - & - & - & 0.389 & 0.111 & 0.047 & 0.888 & 0.982 & 0.996 & 0.814 & 14.19\\
    \checkmark & \checkmark & \checkmark & - & - & - & 0.387 & 0.109 & 0.047 & 0.888 & 0.982 & \textbf{0.997} & 0.817 & 14.01\\
    \checkmark & \checkmark & \checkmark & \checkmark & - & - & 0.381 & 0.108 & 0.046 & 0.894 & 0.984 & \textbf{0.997} & 0.824 & 13.54\\
    \checkmark & \checkmark & \checkmark & - & \checkmark & - & 0.370 & 0.103 & 0.045 & 0.903 & 0.986 & \textbf{0.997} & 0.829 & 13.18\\
    \checkmark & \checkmark & \checkmark & - & \checkmark & \checkmark & \textbf{0.362} & \textbf{0.100} & \textbf{0.043} & \textbf{0.909} & \textbf{0.987} & \textbf{0.997} & \textbf{0.835} & \textbf{12.72}\\
    \bottomrule
    \end{tabular}
    \label{tb:Ablation}
\end{table}
\begin{table}[t]
    \scriptsize
    \addtolength{\tabcolsep}{4.1pt}
    \renewcommand{\arraystretch}{1.1}
    \caption{Effectiveness of the two-phase training scheme.}
    \centering
    \begin{tabular}{l|cccccccc}
    \toprule
    Setting & RMSE & REL & $\log10$ & $\delta_1$ & $\delta_2$ & $\delta_3$ & Kendall's $\tau$ & WHDR(\%)\\
    \midrule
    Single-phase  & 0.386 & 0.107 & 0.046 & 0.892 & 0.984 & \textbf{0.997} & 0.820 & 13.74\\
    Proposed & \textbf{0.362} & \textbf{0.100} & \textbf{0.043} & \textbf{0.909} & \textbf{0.987} & \textbf{0.997} & \textbf{0.835} & \textbf{12.72}\\
    \bottomrule
    \end{tabular}
    \label{tb:two_stage_training}
\end{table}
\begin{table}[h!]
 \begin{minipage}{0.44\linewidth}
    \scriptsize
    \centering
    \addtolength{\tabcolsep}{3pt}
    \renewcommand{\arraystretch}{1.1}
    \caption{Complexity comparison.}
    \begin{tabular}[t]{l|cc}
        \toprule
        & Ranftl~\etal \cite{ranftl2021vision} & Proposed\\\midrule
        \# Params & 130M & 102M  \\
        Speed (fps) & 13.4 & 34.7  \\
        \bottomrule
    \end{tabular}
    \label{tb:complexity_time1}
 \end{minipage} \hspace{0.3cm}
 \begin{minipage}{0.54\linewidth}
    \scriptsize
    \addtolength{\tabcolsep}{2.4pt}
    \renewcommand{\arraystretch}{1.1}
    \caption{Complexity of each component.}
    \centering
    \begin{tabular}[t]{l|ccccc}
        \toprule
        & Encoder & G  & N & M & MDR\\ \midrule
        \# Params & 55M  & 15M & 15M & 15M & 1.7M \\
        Speed (fps) & 50.9 & 475 & 534 & 474 & 447\\
        \bottomrule
    \end{tabular}
    \label{tb:complexity_time2}
 \end{minipage}
\end{table}

{\noindent \bf Ablation studies:}
We conduct ablation studies that add the proposed components one by one in Table~\ref{tb:Ablation}. Here, the 17K images from NYUv2 are used for training. M, N, and G denote the three decoders. MDR$^*$ is the MDR block with $\mu_M$ deactivated. $\dagger$ indicates the use of relative depth data. We see that all components lead to performance improvements, especially in terms of the two relative depth metrics Kendall's $\tau$ and WHDR.

Table~\ref{tb:two_stage_training} shows the effectiveness of the two-phase training scheme of the proposed algorithm. The proposed algorithm, which trains G-Net and N-Net first, shows better results than the single-phase scheme, which trains the entire network at once.

{\noindent \bf Complexities and inference speeds:}
Table~\ref{tb:complexity_time1} compares the complexities of the proposed algorithm and the Ranftl \etal's algorithm~\cite{ranftl2021vision}. The proposed algorithm performs faster with a smaller number of parameters than the Ranftl \etal's algorithm \cite{ranftl2021vision} does. This indicates that the performance gain of the proposed algorithm is not from the increase in complexity but from the effective use of relative depth features. Table~\ref{tb:complexity_time2} lists the complexity of each component of the proposed algorithm. The encoder spends most of the inference time, while the three decoders are relatively fast.

\section{Conclusions}
\label{sec:conclusions}

We proposed a monocular depth estimator that decomposes a metric depth map into a normalized depth map and scale features. The proposed algorithm is composed of a shared encoder with three decoders, G-Net, N-Net, and M-Net, which estimate gradient maps, a normalized depth map, and a metric depth map, respectively. G-Net features are used in N-Net, and N-Net features are used in M-Net. Moreover, we developed the MDR block for M-Net to utilize N-Net features and improve the metric depth estimation performance. Extensive experiments demonstrated that the proposed algorithm provides competitive performance and yields acceptable results even with a small metric depth dataset.

\section*{Acknowledgements}
This work was supported by the National Research Foundation of Korea (NRF) grants funded by the Korea government (MSIT) (No.~NRF-2021R1A4A1031864 and No.~NRF-2022R1A2B5B03002310).
%
%
\bibliographystyle{splncs04}
\bibliography{4247}
\end{document}